\documentclass{article} % For LaTeX2e
\usepackage{iclr2020_conference,times}

\usepackage{amsmath,amsfonts,bm}

\def\eqref#1{equation~\ref{#1}}
\def\1{\bm{1}}

\def\vx{{\bm{x}}}

\DeclareMathAlphabet{\mathsfit}{\encodingdefault}{\sfdefault}{m}{sl}
\SetMathAlphabet{\mathsfit}{bold}{\encodingdefault}{\sfdefault}{bx}{n}

\usepackage{hyperref}
\usepackage{url}
\usepackage{multirow}

\usepackage{booktabs}
\usepackage{amsmath}

\usepackage{caption}
\usepackage{graphicx}

\title{Reject Illegal Inputs with Generative Classifier Derived from Any Discriminative Classifier}

\author{Xin Wang \\ 
Department of Computer Science\\
The University of Hong Kong\\
\texttt{xwang@cs.hku.hk} 
}
\iclrfinalcopy % Uncomment for camera-ready version, but NOT for submission.
\begin{document}

\maketitle

\begin{abstract}
 Generative classifiers have been shown promising to detect illegal inputs including adversarial examples and out-of-distribution samples. Supervised Deep Infomax~(SDIM) is a scalable end-to-end framework to learn generative classifiers. In this paper, we propose a modification of SDIM termed SDIM-\emph{logit}. Instead of training generative classifier from scratch, SDIM-\emph{logit} first takes as input the logits produced any given discriminative classifier, and generate logit representations; then a generative classifier is derived by imposing statistical constraints on logit representations.  SDIM-\emph{logit} could inherit the performance of the discriminative classifier without loss. SDIM-\emph{logit} incurs a negligible number of additional parameters, and can be efficiently trained with base classifiers fixed.  We perform \emph{classification with rejection}, where test samples whose class conditionals are smaller than pre-chosen thresholds will be rejected without predictions. Experiments on illegal inputs, including adversarial examples, samples with common corruptions, and out-of-distribution~(OOD) samples  show that allowed to reject a portion of test samples, SDIM-\emph{logit} significantly improves the performance on the left test sets.
\end{abstract}

\section{Introduction}
Despite the great success of neural network models, they are also surprisingly vulnerable to illegal inputs, like adversarial examples and samples with common corruptions. So model robustness emerge as a big concern. The hardness to achieve  robustness lies in that statistical machine learning models, based on IID assumption, are asked to robustly make predictions on inputs deviated from the training distribution due to adversarial perturbations or corruptions. Considering the unsatisfying situation, an alternative treatment to these deviated inputs is  \emph{rejection without prediction}. Another strong and practical motivation of rejection is the existence of out-of-distribution~(OOD) samples. For example, what should a classifier trained on ImageNet2012 do to a test sample outside of the 1000 training classes? Obviously, a robust classifier should output rejection, while current models give a prediction inside the training classes. 

Supervised Deep Infomax~(SDIM,\cite{wang2020reject}) is an end-to-end framework to learn generative classifiers on data representations, and maximize the mutual information between data and data representations at the same time. SDIMs are able to achieve same-level performance as comparable discriminative classifiers. With SDIM, class conditionals based classification with rejection can be performed to effectively reject illegal inputs, including OOD samples and $\mathcal{L}_p$ adversarial examples. In this paper, we propose SDIM-\emph{logit}, a simple modification of SDIM. SDIM-\emph{logit} is designed to derive a generative classifier given any discriminative classifier. The only difference from SDIM is that SDIM-\emph{logit} takes the logits of the discriminative classifier as inputs, rather than the raw data. The theoretical contribution of SDIM-\emph{logit} is \emph{zero}, but it brings following practical benefits:

\begin{itemize}
    \item SDIM-\emph{logit} enables us to derive generative classifier given any discriminative classifier almost without loss of performance, incurring a negligible number of additional parameters, and training time. We could reuse the many powerful well-trained discriminative models without training from scratch.
    \item We perform classification with rejection in SDIM-\emph{logit}. For each class, one threshold of class conditional is chosen. An input will be rejected if its maximum of class conditionals is smaller than the corresponding threshold.  Illegal inputs, include corrupted inputs, adversarial examples, OOD samples can be effectively rejected. 
\end{itemize}
\paragraph{Difference of Treatment}The big difference between classification with rejection and methods aiming to inherently improve models' recognition robustness lies in the \emph{difference of treatment} to potential illegal inputs. For example, the implicit motivation behind adversarial training is that model should be able to recognize adversarial examples. While for classification with rejection, illegal inputs like adversarial examples are illegal inputs deviated from the training distribution, and should be rejected without prediction. The consequence of attain recognition robustness may be that we have to tackle different illegal inputs separately, and defense mechanisms call for additional designs and training, even independent of the original target model~\citep{song2017pixeldefend,samangouei2018defense}. It is observed that Gaussian noise corruption is highly related to adversarial examples, and   both the adversarial and corruption robustness communities are encouraged to work more
closely together~\citep{ford2019adversarial}. See section~\ref{rw_1} for a more comprehensive list of related works.  The stark advantage of  SDIM-\emph{logit}, also SDIM, is that its classification with rejection as defense mechanism  is a built-in property, and universally applicable to various illegal inputs. 
\iffalse
Following~, two types of robustness we should care about are adversarial robustness and corruption robustness. Both can be defined as functions over the \emph{error set} of a classifier. 
\fi 
\iffalse to answer an orthogonal but complementary problem: could a classifier  reject to make predictions when it has low confidence on test samples.  SDIM-\emph{logit} is a modification of SDIM, 

\paragraph{Connection to Adversarial Robustness and Corruption Robustness}
State-of-the-art image classifiers suffer not only specially crafted adversarial examples, but also
images with common corruptions, like additive Gaussian noises. Unlike adversarial training, which is particularly dedicated to improve the model robustness against adversarial examples. The class conditionals based classification with rejection 

This paper makes \emph{zero} theoretical contribution except for a simple modification of SDIM. However, the modified framework-  

 The number of introduced additional parameters is negligible compared to the discriminative classifier. The training process is very computationally cheap~(taking around 10-20 epochs), and the discriminative classifier is kept fixed. One preferable property of the derived generative classifiers is that we can perform classification with rejection based on the class conditionals. We empirically show \fi 
 
\section{SDIM-\emph{logit}}
\begin{figure*}[hbt!]
    \centering
    \includegraphics[scale=0.4]{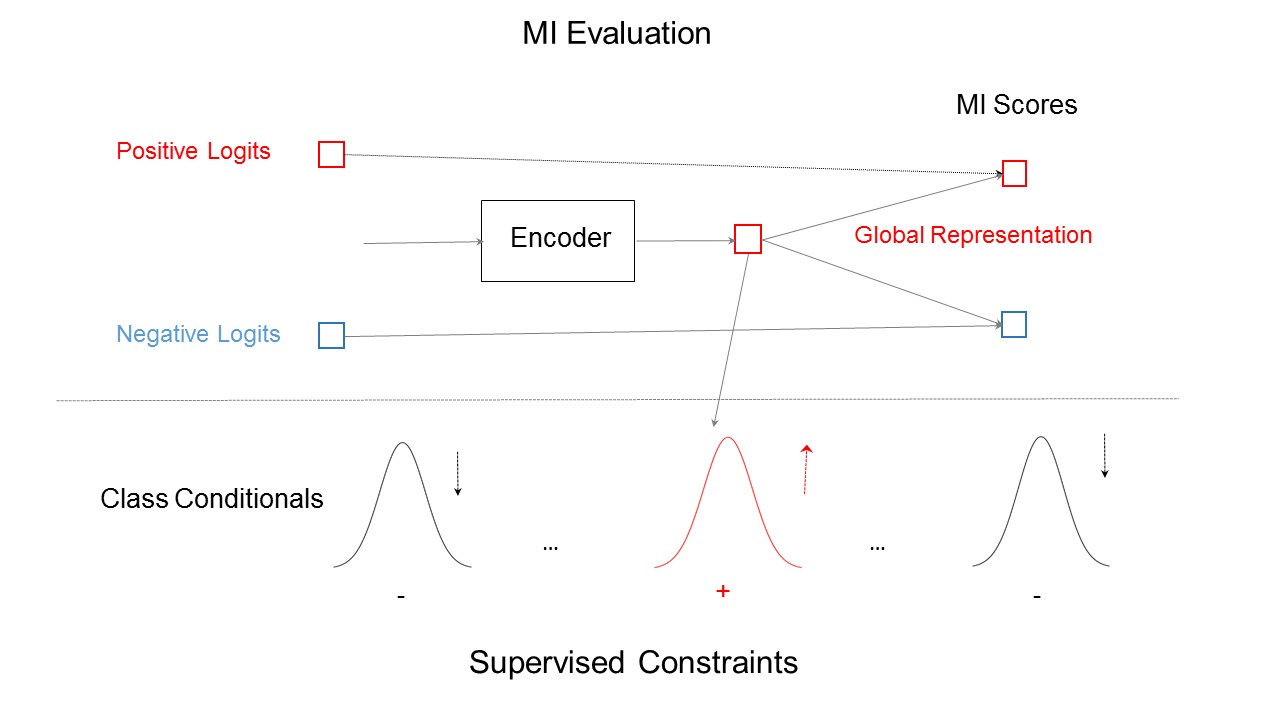}
    \caption{SDIM-\emph{logit} framework. (1)The encoder $E_\phi$ takes as inputs the logits $f(\vx)$, and produces  global representations $E_\phi(f(\vx))$. The MI between logits $f(\vx)$ and global representations $E_\phi(f(\vx))$ is maximized. (2) Supervised constraints are imposed on the global representations $E_\phi(f(\vx))$  for generative classification. }
    \label{fig:sdim_logit}
\end{figure*}

Let $f(x)$ denote logits (before softmax) of a discriminative classifier. SDIM-\emph{logit} consists of three components: logit encoder $E_\phi$, MI evaluation network $T_\omega$, and $C$-way embedding layer of Gaussian class conditionals. The framework is summarized in Fig.~\ref{fig:sdim_logit}.

\paragraph{Maximize MI}The MI evaluation network $T_{\omega}$ maps positive pairs and negative pairs to MI scores specified by corresponding MI lower-bound. Negative pairs are simply obtained by combine all unpaired ones within the same mini-batch. Equivalently, we minimize the following loss $\mathcal{J}_{\text{MI}}$:
\begin{equation}
    \mathcal{J}_{\text{MI}} = - \Tilde{\mathcal{I}} (f(\vx), E(f(\vx))),
\end{equation}
where $\Tilde{\mathcal{I}}$ is some specific lower-bound of MI~(see \cite{wang2020reject,hjelm2018learning} for more details).

\paragraph{Generative Classification} Supervised constraints are imposed on the global representations $\Tilde{\vx} = E_{\phi}(\vx)$ with loss $\mathcal{J}_{\text{NLL}} + \mathcal{J}_{\text{LM}}$ for generative classification~\citep{wang2020reject}. The true class conditionals  are maximized, while false class conditionals are minimized. 

\iffalse
\paragraph{Maximizing the Class Conditionals} Similar to SDIM, the class conditionals are evaluated with a class-wise Gaussian embedding layer on the logits, denoted as $\Tilde{\vx}$ of input $\vx$.

SDIM-\emph{logit} allows to make use of the 
\begin{equation}
    \mathcal{J}_{\text{CE}}= \text{cross\_entropy}\big (\text{softmax}(\frac{\log p(\Tilde{\vx}|y)}{\tau}), y \big),
    \label{eq:ce}
\end{equation} where $\tau$ is a temperature factor.
Eq.(\ref{eq:ce}) is equivalent to maximize the true class conditional $\log p(\Tilde{\vx}|y)$. Using a higher temperature $\tau$ encourages larger gaps between the true and false class conditionals, i.e. minimizing the false class conditionals. \fi 

We end up with minimizing the following loss:
\begin{equation}
    \mathcal{J}_{\text{SDIM-\emph{logit}}} = \alpha \cdot \mathcal{J}_{\text{MI}} + \beta \cdot \mathcal{J}_{\text{NLL}} + \gamma \cdot \mathcal{J}_{\text{LM}},
    \label{eq:loss}
\end{equation}
where $\alpha, \beta, \gamma$ are scaling factors.

\paragraph{Classification with Rejection} Normally, the classifier should  output the class label $y^*$ that maximizes class conditional probability of test sample $\vx^*$. Here,  we add a rejection option by setting a threshold for each class conditional probability distribution, and define our decision function with rejection as:
\begin{equation}
    \begin{cases}
      y^*, & \text{if } \log p(\vx^*| y^*) \geq \delta_{y^*}\\
      \textit{Rejection}, & \text{otherwise}
    \end{cases}
\end{equation}
The model gives a rejection when $\log p(\vx^*| y^*)$ is smaller than the threshold $\delta_{y^*}$ of class $y^*$.
\section{Related Works}
\subsection{Improving Robustness On Various Illegal Inputs}
\label{rw_1}
\paragraph{Adversarial Examples} Adversarial training improves the adversarial robustness of victim models, but could incur even more computations than traditional training due to the necessity to generate adversarial examples. A line of works aim to accelerate adversarial training~\citep{wong2020fast, qin2019adversarial}.  Adversarial training can result in performance degradation on clean test sets. Recently, AdvProp~\citep{xie2019adversarial} first solve this problem by using a separate batch norm for adversarial examples, considering the underlying distribution of them is already deviated from the normal training one. 
\iffalse \paragraph{Samples with Common Corruptions} \fi 
\paragraph{OOD Samples} \cite{ren2019ratio} find that the likelihood of an image  is dominated by the irrelevant background statistics.  They propose to train a background model, and correct with the ratio between original likelihood and background likelihood. These works tend to solve one type of illegal inputs individually, and not applicable to others.

\iffalse
 Many efforts have been put to build more robust classifiers that are able to classify various adversarial examples. Adversarial training probably is the most effective way to improve the models' performance on $\mathcal{L}_p$ adversarial examples.  All these works are complementary and orthogonal to SDIM-\emph{logit}, which outputs rejection of a prediction request when it has low confidence to make correct prediction on it.   \fi 

\subsection{generative modeling for classification robustness}
\paragraph{Generative Classifiers on Raw Data} Deep generative models have achieved great successes in synthesising realistic samples. So they can be great candidates to model the class conditional probabilities $p(\vx|y)$ required for generative classifiers. However, it is far from achieving  comparable classification performance to the discriminative counterparts.\cite{li2018generative,schott2018towards,ethan2019conditional} explore the adversarial robustness of fully generative classifiers, they all only get satisfying accuracy on MNIST and hardly achieve acceptable classification performance even on CIFAR10. They can demonstrate counter-intuitive behaviours, and assign surprisingly higher likelihoods to OOD samples~\citep{nalisnick2018deep,choi2018generative}.

\paragraph{Generative Classifiers on Data Representations}
 \cite{lee2020deep} propose deep learning objective to learn the multi-class generative classifier, by
fusing the concept of Gaussian discriminant analysis with DNNs directly. It shares some similarity to the supervised constraints of SDIM-\emph{logit}, but SDIM-\emph{logit} ensure the quality of learned representations by maximizing $\mathcal{J}_{\text{MI}}$. Their evaluations are particularly on OOD detection, while ours are on various illegal inputs that may deviate from training distribution.

\section{Experiments}
In our experiments, we choose ResNet18~\citep{he2016deep} and ResNeXt-29 $4\times 64d$~\citep{xie2016aggregated} as the base discriminative classifiers. The code~\footnote{Code of SDIM-\emph{logit}: \href{https://github.com/wangxin0716/SDIM_logits}{https://github.com/wangxin0716/SDIM\_logits.}} is open sourced  for reproduction.  Then we derive generative classifiers from the base classifiers with SDIM-\emph{logit}.  SDIM-logit introduces 12k additional parameters, which is negligible compared to the base model, which has 21M parameters. During the training, the base models are fixed, and only the introduced parameters of SDIM-\emph{logit} are trainable. The experiments aim to show that: (1) the training of SDIM-\emph{logit} is computationally efficient, and the derived generative classifiers inherit the performance of base classifiers without observed losses; (2) We perform classification with rejection based on class conditionals of derived generative classifiers. 

\iffalse
We evaluate SDIM-\emph{logit} on three robustness benchmark test sets: 
\begin{itemize}
    \item ImageNet-A~\citep{hendrycks2019nae} consists of 7500 unmodified, naturally occurring but difficult images for even powerful classifiers. For example, DenseNet-121~\citep{huang2017densely} obtains around $75\%$ test accuracy on ImageNet, but only around $2\%$ accuracy on ImageNet-A.
     \item ImageNet-C~\citep{hendrycks2019robustness}:  
      \item ImageNet-P~\citep{hendrycks2019robustness}:  
\end{itemize}
The goal is to show whether SDIM-\emph{logit} can effectively reject the prediction requests when it has low confidence on them. Evaluations on $\mathcal{L}_p$ norm ball adversarial examples are not covered in this paper, and is left in future work. \fi 

\paragraph{Efficient Training}
 All SDIM-\emph{logit} models in our experiments are trained in 40 epochs using 1 Nvidia 1080 Ti GPU, and it takes only 20s for ResNet18 and 80s for ResNeXt-29 per epoch. 

\paragraph{Choice of thresholds} Following~\citep{wang2020reject}, for each class, we choose to evaluate on two different thresholds: \emph{1st} and \emph{2nd} percentiles of class conditional log-likelihoods of the correctly classified training samples. 
\subsection{Performance on Clean Test Sets}
All the inputs are scaled in [0, 1] without normalizations, which we think is the reason that the results reported here are slight worse than that in the original papers. Similar results to \cite{wang2020reject} are obtained that a rejection option help reject low-confidence test samples. Using a higher threshold, which also implicitly rejects some legal test samples,  leads to higher accuracy on the left test sets.
\begin{table}[!h]
\footnotesize
\begin{center}
\begin{tabular}{l c c c c c c}
\toprule
\multirow{2}{*}{Base Classifier} & \multirow{2}{*}{Base Acc.} & \multirow{2}{*}{SDIM-\emph{logit} Acc.} & \multicolumn{2}{c}{$1st$ Percentile } &  \multicolumn{2}{c}{$2nd$ Percentile }\\
%\cline{3-6}
& & & Acc. Left & Rej. Rate & Acc. Left & Rej. Rate\\
 \midrule
ResNet-$18$ & 95.16\% & 95.01\% & 95.64\% & 4.10\% & 96.19\%& 6.28\% \\
ResNeXt-29 $4 \times 64d$ & 95.81\% & 95.24\% & 97.76\% & 4.19\% & 98.04\% & 5.85\%\\
\bottomrule
\end{tabular}
\end{center}
\caption{Classification performances of SDIM-\emph{logit} models. The second and third columns are the accuracy of base models and derived  models with SDIM-\emph{logit}. For implications of  using the proposed decision function with rejection, we report the rejection rates of the test sets and the classification accuracy on the left test sets for the chosen thresholds. }
\label{tab:acc}
\end{table}

\subsection{Evaluations on Corrupted Dataset CIFAR10-C}
CIFAR10-C~\citep{hendrycks2019robustness} \footnote{CIFAR10-C download URL: \url{https://zenodo.org/record/2535967}} is a benchmark dataset based  on CIFAR10 with 15 diverse common corruptions. Each corruption has 5 severity levels. These corruptions cause samples to deviate from the training distribution. Results in Tab.~\ref{tab:cifar10_c_results} show that with the increasing of severity level, models' performance degrade and increasing portion of test samples will be rejected. In general, our rejection decision function improves models' accuracy  by around $20\%$. 
\begin{table}[!ht]
\footnotesize
\begin{center}
\begin{tabular}{l c c c c c c}
\toprule
SDIM-\emph{logit} & \multirow{2}{*}{Severity} & \multirow{2}{*}{Acc. w/o Rej.~(\%)} & \multicolumn{2}{c}{$1st$ Percentile~(\%) } &  \multicolumn{2}{c}{$2nd$ Percentile~(\%) }\\
 Base Model&  & & Acc. Left & Rej. Rate & Acc. Left & Rej. Rate\\
 \midrule
\multirow{6}{*}{ResNeXt-29 4$\times$ 64d} & 1 & 87.60 & 97.85 & 29.87 & 98.11 & 32.44\\
 & 2 & 80.79 & 96.58 & 40.32 & 97.00 & 43.00\\
 & 3 & 73.96 & 95.43 & 48.80 & 96.09 & 51.42\\
 & 4 & 65.97 & 92.70 & 57.05 & 93.81 & 59.56\\
 & 5 & 54.10 & 84.67 & 67.56 & 86.46 & 69.95\\
\cmidrule(l){2-7}
 &  mean & 72.43 & 93.45 & 48.72 & 94.30 & 51.27\\
\midrule
\multirow{6}{*}{ResNet18} & 1 & 87.73 & 96.58  & 23.47 & 97.07 & 26.43\\
 & 2 & 81.61 & 94.07 & 32.23 & 94.80 & 35.70\\
 & 3 & 75.33 & 91.03 & 39.88 & 92.08 & 43.57\\
 & 4 & 67.74 & 86.49 & 47.95 & 87.91 & 51.83\\
 & 5 & 56.04 & 77.73 & 59.08 & 79.77 & 63.03\\
 \cmidrule(l){2-7}
 & mean & 73.69 & 89.18 & 40.52 & 90.32 & 44.11\\
\bottomrule
\end{tabular}
\end{center}
\caption{Classification performance of the generative classifiers derived with SDIM-\emph{logit} on CIFAR10-C. Detailed results with different severity levels~(1-5, and \emph{mean}), and with different decision functions are reported. The third column shows the classification accuracy of original decision function, while the latter columns show the rejection rates and accuracies on the left test sets of our rejection function with different thresholds.}
\label{tab:cifar10_c_results}
\end{table}

\subsection{Evaluations on  Adversarial Examples}
In this section, we demonstrate the effectiveness of the rejection on  adversarial examples generated by $\mathcal{L}_{\infty}$ Projected Gradient Descent(PGD,~\cite{madry2017towards}).  With the increasing of epsilon~(infinite norm budget of adversarial perturbations),  SDIM-\emph{logit} models achieve higher rejection rates of  adversarial examples.  In \citep{li2018generative}, they use fully generative models to model the class conditionals;  For its strongest variant GBZ (see its Fig. 4), all the three proposed detection methods report detection rates $ < 50\%$ on even CIFAR10-binary. Note that their models can not achieve acceptable performance on CIFAR10.

\iffalse In the generation of adversarial examples, we treat the derived generative classifiers the same way as the discriminative classifiers, except that the logits of SDIM-\emph{logit} are log class conditionals on the data representations $\log p(\Tilde{\vx}|y)$. 

 Furthermore, we also show that classification with rejection is complementary to methods that inherently improve the adversarial robustness. We also train base classifiers with Gaussian noise augmentation to improve adversarial robustness~\citep{ford2019adversarial}. \fi 

\begin{figure*}[hbt!]
    \centering
    \includegraphics[scale=0.36]{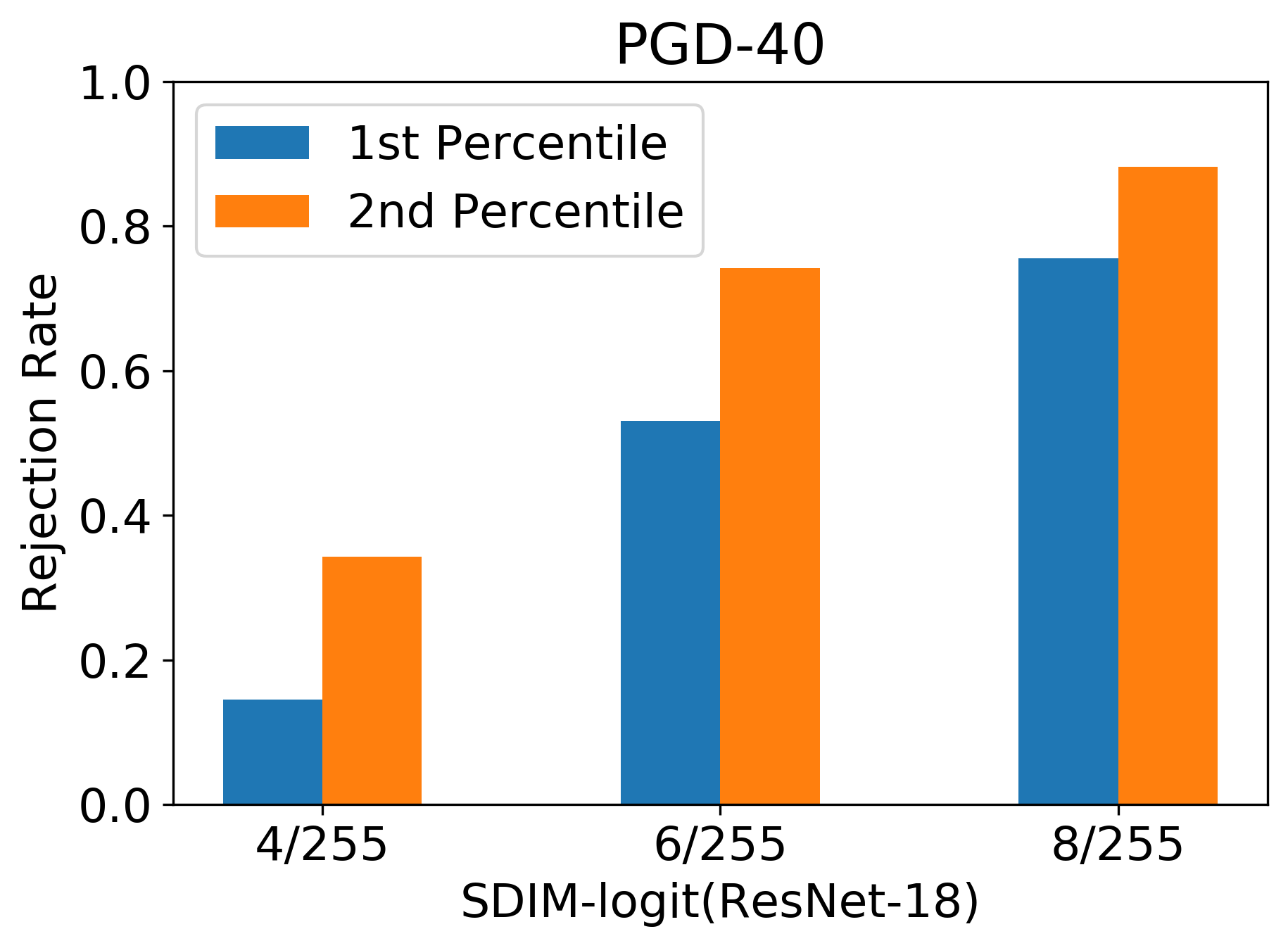}
    \hspace{0.5cm}
    \includegraphics[scale=0.36]{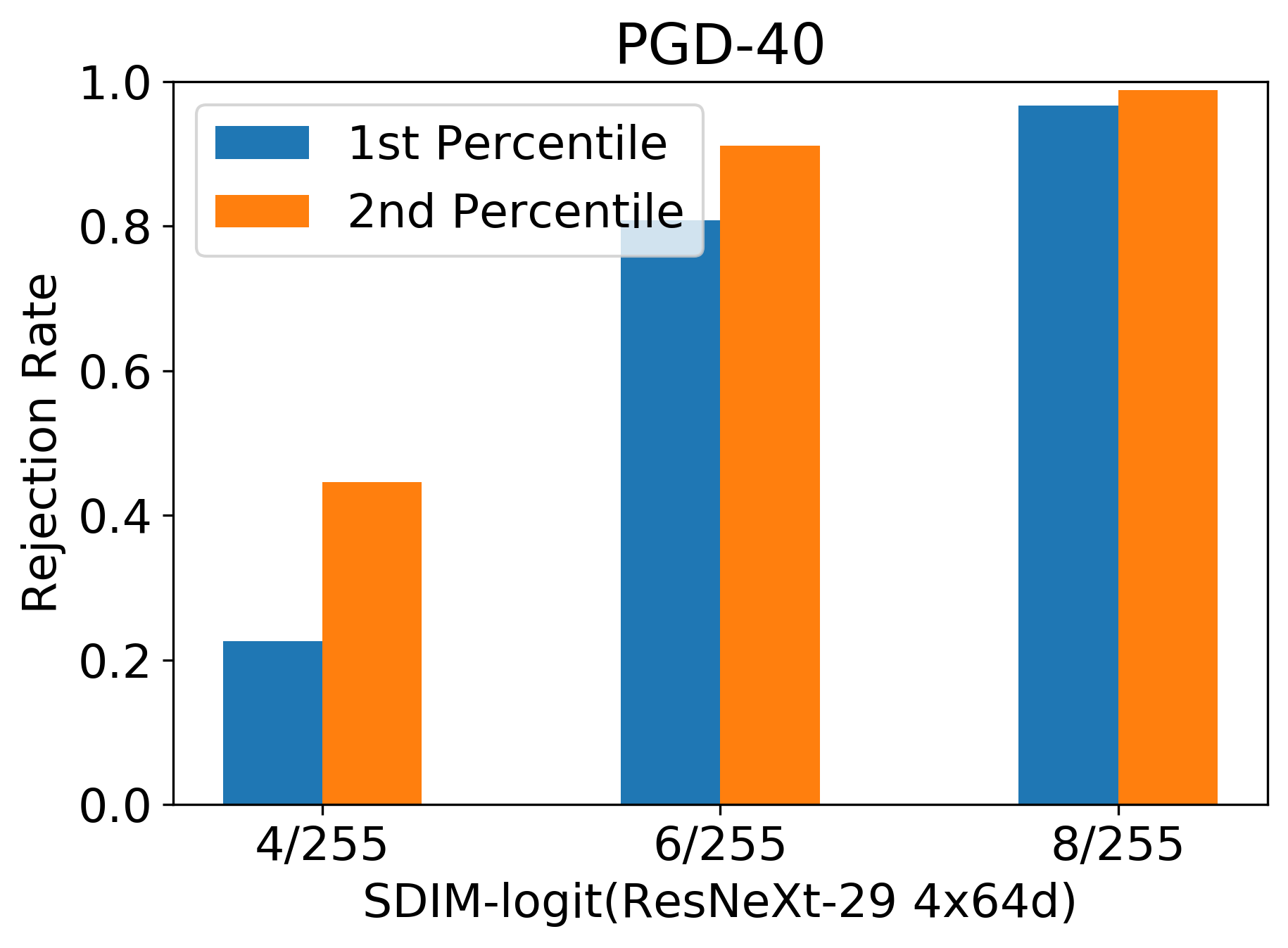}
    \caption{Rejection~(detection) rates of derived generative classifiers with rejection. Evaluations are on $\mathcal{L}_{\text{inf}}$-PGD with epsilons $\epsilon =\{4/255, 6/255, 8/255\}$. The number of iterations is 40, and the step size is 0.01~(same as \citep{li2018generative,madry2017towards}).  }
    \label{fig:adv_detection}
\end{figure*}
\subsection{OOD Detection}
We evaluate SDIM-\emph{logit}'s performance to detect OOD samples on the typical distribution pair: CIFAR10(in)-SVHN(out), on which fully generative models like VAEs and Flows fail~\citep{nalisnick2018deep,choi2018generative}. We report better results than SDIM~(see Tab.~\ref{tab:ood_detection}). This is reasonable since on clean CIFAR10 test set, our SDIM-\emph{logit} models($>95\%$) perform better than SDIM($\sim 92\%$).

\begin{table}[!h]
\footnotesize
\begin{center}
\begin{tabular}{l c c }
\toprule
\multirow{2}{*}{Model}  & \multicolumn{2}{c}{Rejection Rate~(\%)} \\
& 1st Percentile & 2nd Percentile\\
 \midrule
SDIM      & 94.24  & 95.81 \\
SDIM-\emph{logit}(ResNet-18)      & 98.40 & 98.75 \\
SDIM-\emph{logit}(ResNeXt-29 $4\times 64d$)      & 98.54 & 98.77 \\
\bottomrule
\end{tabular}
\end{center}
\caption{ OOD detection results of SDIM-\emph{logit} and SDIM on CIFAR10-SVHN distribution pair.}
\label{tab:ood_detection}
\end{table}
\iffalse
\begin{table}[!h]
\footnotesize
\begin{center}
\begin{tabular}{l c c c c c c c c}
\toprule
SDIM-\emph{logit} & \multicolumn{2}{c}{FGSM} & \multicolumn{2}{c}{PGD} & \multicolumn{2}{c}{DeepFool} &  \multicolumn{2}{c}{CW}\\
 Base Model  & 1\emph{st} Per. & 2\emph{nd} Per.& 1\emph{st} Per. & 2\emph{nd} Per.& 1\emph{st} Per. & 2\emph{nd} Per.& 1\emph{st} Per. & 2\emph{nd} Per.\\
 \midrule
 \footnotesize{w/o  \emph{Adversarial Training}}\\
ResNeXt & 1 & 86.04 & 96.99 & 28.58 & 97.33 & 31.02\\
ResNet & 1 & 87.80 & 97.06  & 24.57 & 97.33 & 27.14\\
\midrule
\footnotesize{w/ \emph{Adversarial Training}}\\
ResNeXt & 1 & 86.04 & 96.99 & 28.58 & 97.33 & 31.02\\
ResNet & 1 & 87.80 & 97.06  & 24.57 & 97.33 & 27.14\\
\bottomrule
\end{tabular}
\end{center}
\caption{Detection results of derived generative classifiers on PGD and CW attacks. Results on base classifiers with and without adversarial training are reported.}
\label{tab:adv_example_results}
\end{table}
\fi 
\section{Conclusions}
Image recognition models are trained to recognize samples of limited number of classes, thus are no way supposed to always make predictions on all inputs in unlimited space. Being able to reject what the models do not know is important for practical deployment. In this paper, we propose SDIM-\emph{logit}, which derive generative classifier given any discriminative classifier without loss of performance. The training of SDIM-\emph{logit} is computationally cheap, and the base discriminative classifier is fixed. With SDIM-\emph{logit}, classification with rejection can be performed to reject various illegal inputs, including adversarial examples, corrupted inputs, and OOD samples. One thing to note is that classification with rejection is complementary and orthogonal to other defense methods aiming to improve model recognition robustness. It is promising to build more robust models by  exploring their combinations. For example, though adversarial training effectively improves adversarial robustness, the adversarial accuracy so far is still not satisfying. Thus introducing a rejection option is necessary to reduce the threaten of adversarial examples.

\bibliography{iclr2020_conference}
\bibliographystyle{iclr2020_conference}

\iffalse
\appendix

\section{Appendix}
You may include other additional sections here. 
\fi
\end{document}